\documentclass[10pt,journal]{IEEEtran}

\usepackage{amsmath,amssymb}
\usepackage{graphicx}
\usepackage{booktabs}
\usepackage{array}
\usepackage[hidelinks]{hyperref}
\usepackage{microtype}
\usepackage{newunicodechar}

\graphicspath{{../figures/}}

\newunicodechar{—}{, }
\newunicodechar{–}{--}
\newunicodechar{−}{\ensuremath{-}}
\newunicodechar{→}{\ensuremath{\rightarrow}}
\newunicodechar{Δ}{\ensuremath{\Delta}}
\newunicodechar{±}{\ensuremath{\pm}}
\newunicodechar{×}{\ensuremath{\times}}
\newunicodechar{…}{\ldots}
\newunicodechar{∈}{\ensuremath{\in}}
\newunicodechar{≤}{\ensuremath{\leq}}
\newunicodechar{≥}{\ensuremath{\geq}}
\newunicodechar{≈}{\ensuremath{\approx}}
\newunicodechar{≫}{\ensuremath{\gg}}
\newunicodechar{·}{\textperiodcentered}
\newunicodechar{μ}{\ensuremath{\mu}}
\newunicodechar{✓}{\checkmark}
\newunicodechar{✗}{\ensuremath{\times}}

\begin{document}

\title{Different Teachers, Different Capabilities: Sub-1B On-Device Distillation for Structured Text Enrichment}

\author{Vinay~Kumar~Chaganti
\thanks{The author is an independent researcher (e-mail: cvk.atreya@gmail.com).}%
\thanks{Evaluation-harness code, figure-generation scripts, and portions of the manuscript were produced with AI-agent assistance; see the Acknowledgment.}}

\markboth{Preprint, July 2026}{Chaganti: Different Teachers, Different Capabilities in Sub-1B Structured-Output Distillation}

\maketitle

\begin{abstract}
High-volume structured extraction pays a large model's latency on every item, so distilling the task into a small on-device model is attractive: comparable output at a fraction of the time and cost. We measure what that distillation actually delivers, per sub-task. Each news article is mapped to one JSON object with a short summary and five categorical labels. We distill an 8B reasoning teacher (\texttt{deepseek-r1:8b}) into a 0.6B student (Qwen3-0.6B; QLoRA, three seeds), and add two teacher controls: a same-size non-reasoning teacher and a larger managed pipeline. A blinded, reference-free, three-judge panel scores every arm against the full article, alongside two non-distillation baselines, few-shot prompting and constrained decoding. The student runs at about 0.8 s per article against the teacher's 39 s, and recovers 58\% of the base-to-teacher gap on summary quality, beating its primary baseline (constrained decoding) by +16.8 points and few-shot prompting by a secondary +4.9. A same-size non-reasoning teacher trains a student no better than the untuned base, so the summary gain follows from the teacher's reasoning nature rather than its scale. Capabilities then split by teacher: the reasoning teacher transfers writing quality and the managed pipeline transfers label diversity, while a same-size instruction teacher's students stay more grounded on the 22 short, thin-source articles in the 93-item test set (74 versus 55 faithful), where the reasoning-lineage student fabricates. That grounding difference is a consistent ordering rather than a significant aggregate effect, and the subgroup is small, so we report it as a direction. Because no single engine wins every field, the deliverable is a per-field routing map for on-device enrichment.
\end{abstract}


\section{Introduction}\label{sec:intro}
\IEEEPARstart{H}{igh}-volume structured extraction is a common production pattern: one prompt, one schema-bound JSON object per item, repeated over thousands of items. It appears in ticket triage, log classification, document intake, moderation pre-filters, and catalog extraction. Run through a mid-sized or large model, it pays that model's latency on every item. Distilling the task into a small on-device model promises comparable output far faster and at lower cost. This paper measures what that distillation delivers, per sub-task, against the cheaper alternatives a practitioner tries first.

The instance we study is news enrichment. Each article is mapped to one JSON object with a short summary and five closed-vocabulary labels: \emph{sentiment}, \emph{urgency}, \emph{frame}, \emph{tone}, and \emph{depth}. An 8B reasoning teacher (\texttt{deepseek-r1:8b}) produces acceptable output but takes about 39 s per article, so a 500-item batch runs 5.4 hours on a consumer laptop. A 0.6B model is roughly 40 times faster and fits on-device; the open question is quality. Distillation raises a small model's output toward the teacher's, and we ask how much of that quality it recovers on each sub-task, and whether it preserves faithfulness, the one property a user-facing pipeline cannot compromise. We report per sub-task because the sub-tasks carry different failure costs: a single aggregate hides the axis where a regression is unacceptable.

Every model under study runs locally. Fixed weights at temperature 0 give deterministic, unlimited reruns, so the measurement holds still and every reported number recomputes offline. Only the judge panel and the larger third teacher are hosted.

The findings are per-field, not a single verdict. Distillation recovers most of the teacher's summary quality and beats both non-distillation baselines. A same-size non-reasoning-teacher control shows this gain follows from the teacher's reasoning nature rather than its scale: an equal-size non-reasoning teacher trains a student no better than the untuned base. Capabilities then split by teacher. The reasoning teacher transfers writing quality; a larger managed pipeline transfers label diversity; and a same-size instruction teacher's students stay more grounded on thin sources, where the reasoning students fabricate, though this last effect is a consistent ordering rather than a significant aggregate difference. Short articles are hard for the student across nearly every quality axis, and the thin-source faithfulness finding rests on only 22 of the 93 test articles, so we treat it as a direction. The deliverable is a per-field routing map, not a blanket recommendation to use the small model.

The evaluation is built for robustness. It uses 93 held-out articles; twelve arms across three distillation teachers, two non-distillation baselines, and the untuned base; three training seeds; a family-independent three-judge panel validated by a negative control; and paired-bootstrap significance tests.

Our contributions, findings first, are: (1) a same-size non-reasoning-teacher control that separates the teacher's reasoning nature from its scale and shows the summary gain depends on the former and does not extend to faithfulness; (2) a capability split across three teachers, in which reasoning transfers writing quality and scale with synthetic data transfers label diversity, with the same-size instruction teacher holding thin-source grounding as a consistent but non-significant ordering, none of which an aggregate score reveals; (3) a per-field routing map for on-device enrichment; and (4) a fully local, reference-free, human-free evaluation harness (a decomposed binary checklist, gap-closure-to-teacher reporting, two non-distillation baselines, a negative control, and a full-source faithfulness-grading result) whose metrics reproduce offline from a released scorecard.

We state the bounds plainly: there are no human gold labels; a three-judge panel does not make magnitudes judge-invariant; and classification accuracy is measured against panel consensus, not ground truth. Section~\ref{sec:limits} collects these.

\section{Related Work}\label{sec:related}
\textbf{Distilling into small on-device models.} Training a small student to imitate a larger teacher dates to Hinton \emph{et al.} \cite{hinton2015}, with sequence-level distillation for generation from Kim and Rush \cite{kimrush2016}; the modern black-box variant supervises a small model on a large model's outputs \cite{xusurvey2024, zhu2023}. Sub-billion on-device models are now a deliberate design target \cite{yang2025, liu2024mobilellm}, and distillation is a standard route to their capability \cite{rang2025}. Closest to this work are targeted distillations for structured extraction: UniversalNER distills an LLM into a small open-NER student \cite{zhou2023universalner}, and a distilled biomedical extractor beats its own teacher at a fraction of the size \cite{gu2023biomedical}. Most such work reports one aggregate quality delta; this study decomposes by sub-task and adds non-distillation controls, so distillation is credited only for what cheaper levers do not already provide.

\textbf{Which teacher, and does reasoning help?} DeepSeek-R1 \cite{deepseek2025} showed reasoning ability transfers to smaller dense models, triggering a wave of replications \cite{zhang2025, li2025} concentrated on math, code, and reasoning benchmarks at 1.5B--32B scale, not the non-reasoning production tasks studied here. Whether a stronger or reasoning teacher helps is contested. Controlled studies find teacher task-fit, not benchmark score, governs student gains \cite{chen2025factors}; that distillation improves a student's outputs without transferring the teacher's fidelity \cite{ramesh2025}; that teacher size has an \emph{optimum} rather than a maximum \cite{busbridge2025}; and that too large a teacher--student gap degrades the student \cite{mirzadeh2020}. On simple inputs reasoning can actively hurt: overthinking and chain-of-thought overhead transfer through distillation \cite{chen2024overthink, zheng2025cot, aggarwal2025}, and reasoning-oriented training can \emph{raise} hallucination \cite{yao2025}. This study trains on the teacher's final answers only (thinking disabled), the deployment-realistic recipe; rationale supervision \cite{hsieh2023} is left to future work. To our knowledge no prior work isolates the teacher's reasoning nature with a same-size non-reasoning-teacher control on a non-reasoning structured task, which is what Section~\ref{sec:reasoning} does.

\textbf{Synthetic-data distillation.} The managed-pipeline arm expands its training data synthetically, a lineage from Self-Instruct \cite{wang2023selfinstruct} to recent work isolating synthetic data's contribution within distillation \cite{shirgaonkar2024}.

\textbf{Faithfulness in small and compressed summarizers.} That the smaller, non-reasoning lineage is the most grounded here echoes a counterintuitive prior result: pruned models can hallucinate \emph{less}, leaning harder on the source \cite{chrysostomou2023}. Faithfulness also varies with input characteristics \cite{ramprasad2024} and tracks instruction-tuning more than scale \cite{zhang2023newssumm}, consistent with the thin-source and teacher-nature effects reported below. On-device serving quantizes the model \cite{song2025ondevice, lin2023awq}; every arm here is measured at its deployed 4-bit quantization.

\textbf{LLM-as-a-judge and panels.} Scalar LLM judges suffer position, verbosity, and self-preference biases and saturate near the ceiling \cite{zheng2023mtbench, wang2023unfair, ye2024bias, liu2023}. Two responses shape this harness. Decomposition: binary checklists and decomposed-requirement pass-rates are more reliable than a 0--5 score \cite{lee2024, cook2024tick, qin2024infobench}. And panels over single judges: a panel of diverse models reduces intra-model bias and tracks humans better than one large judge \cite{verga2024poll}, and an ensemble is better calibrated than an overconfident single judge \cite{tian2025overconfidence}. Because evaluators favor their own and same-family generations \cite{panickssery2024, wataoka2024}, the panel here excludes the teacher's and student's families. Faithfulness is graded reference-free against the source, related to atomic-fact and entailment checkers \cite{min2023, tang2024} and to reference-free metrics that score against the input rather than a gold reference \cite{jiang2023tigerscore}, and distinct from consistency-based detection \cite{manakul2023}, which uses no source. Schema-constrained-generation benchmarks test JSON validity \cite{geng2025jsonschema} but not content correctness, the gap this content grading fills.

\textbf{What is being distilled.} \texttt{deepseek-r1:8b} is DeepSeek-R1-Distill-Llama-8B, itself distilled from the 671B R1 \cite{deepseek2025}, and Qwen3-0.6B is itself a distillation product \cite{yang2025}. The pipeline is therefore third-order distillation, which we conjecture contributes to the achievable ceiling, though the 0.6B student's own capacity is an equally plausible binding constraint (Section~\ref{sec:limits}).

\section{The Structured Enrichment Task}\label{sec:task}
\begin{figure}[t]\centering
\includegraphics[width=\columnwidth]{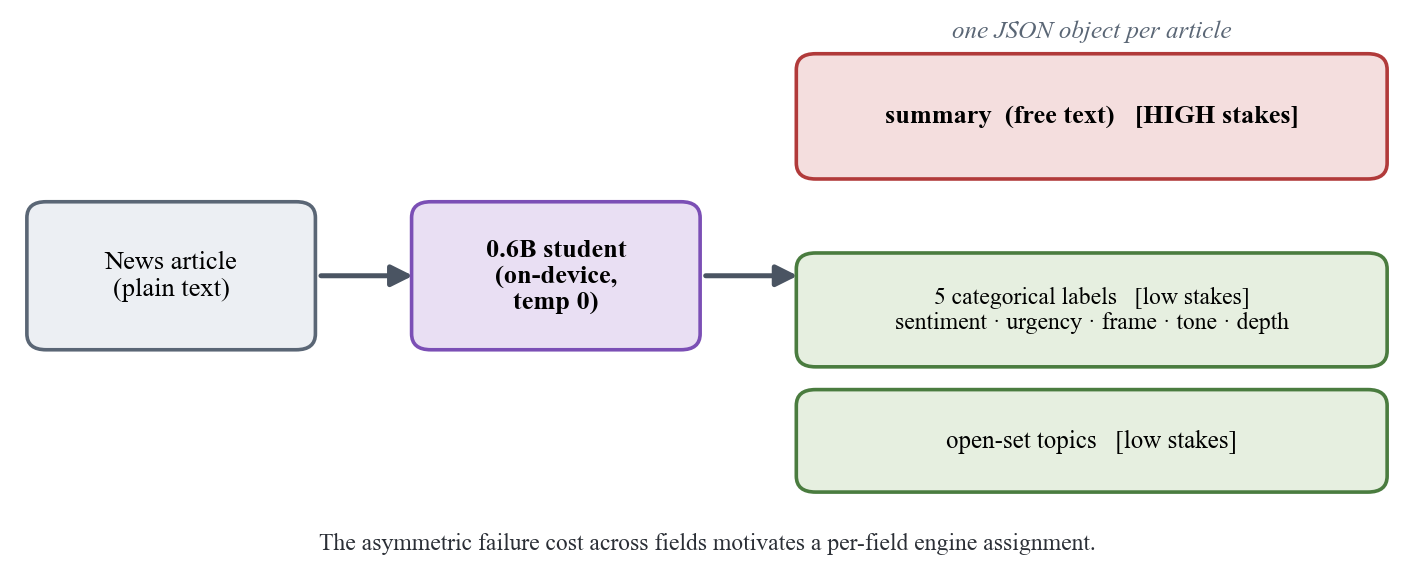}
\caption{The structured-enrichment task. Each article maps to one JSON object, a free-text summary, five categorical labels, and open-set topics. The failure costs are asymmetric: a hallucinated summary misleads, while a wrong label or topic is cheap. That asymmetry motivates the per-field engine assignment (Section~\ref{sec:assignment}).}
\label{fig:panel}
\end{figure}

Each article is enriched with one JSON object of seven fields: a free-text summary, five categorical labels, and an open-set topic list (Fig.~\ref{fig:panel}). A representative teacher output pairs a three-to-four-sentence summary with the label fields, e.g., \emph{sentiment}=positive, \emph{urgency}=developing, \emph{frame}=analytical, \emph{tone}=analytical, \emph{depth}=standard, and \emph{topics}=\{AI~Safety, Responsible~AI\}.

The categorical fields draw from small closed vocabularies observed in the teacher's own labels: \emph{sentiment} $\in$ \{positive, neutral, negative\}; \emph{urgency} $\in$ \{breaking, developing, evergreen\}; \emph{frame} $\in$ \{analytical, conflict, human\_interest, economic\}; \emph{tone} $\in$ \{analytical, optimistic, opinion, alarming\}; \emph{depth} $\in$ \{brief, standard, deep\_dive\}. Several vocabularies are imbalanced in the teacher's own labels (\emph{depth}, for instance, is 71\% \emph{standard} with \emph{deep\_dive} near-absent; full distributions in Table~\ref{tab:corpus}), which matters directly for interpreting per-field accuracy and is made concrete with a majority-class baseline in Section~\ref{sec:classification}.

This output is really two machine-learning problems with different success criteria (Table~\ref{tab:subtasks}), free-text summarization and closed-set classification, and scoring them together hides what distillation moved. Open-set topics are reported as one coverage check inside the summary rubric, not a third sub-task, since there is no per-topic gold to score against. The asymmetry in failure cost is why the study's conclusion is a per-field engine assignment (Section~\ref{sec:assignment}) rather than a single verdict: the quality bar a field must clear depends on what a wrong value costs.

\begin{table}[t]
\centering
\caption{The Enrichment Output Is Two Sub-Tasks With Different Failure Costs}
\label{tab:subtasks}
\small
\setlength{\tabcolsep}{4pt}\begin{tabular}{@{}p{2.2cm}p{1.8cm}p{2.6cm}@{}}
\toprule
Sub-task & Type & Failure cost downstream \\
\midrule
Summarization (\texttt{summary}) & free-text generation & high, a hallucinated summary misleads \\
Classification (5 labels) & closed-set labels & low, a mislabeled field is cheap \\
\bottomrule
\end{tabular}
\end{table}

\section{Experimental Setup}\label{sec:setup}
\begin{figure*}[t]\centering
\includegraphics[width=0.92\textwidth]{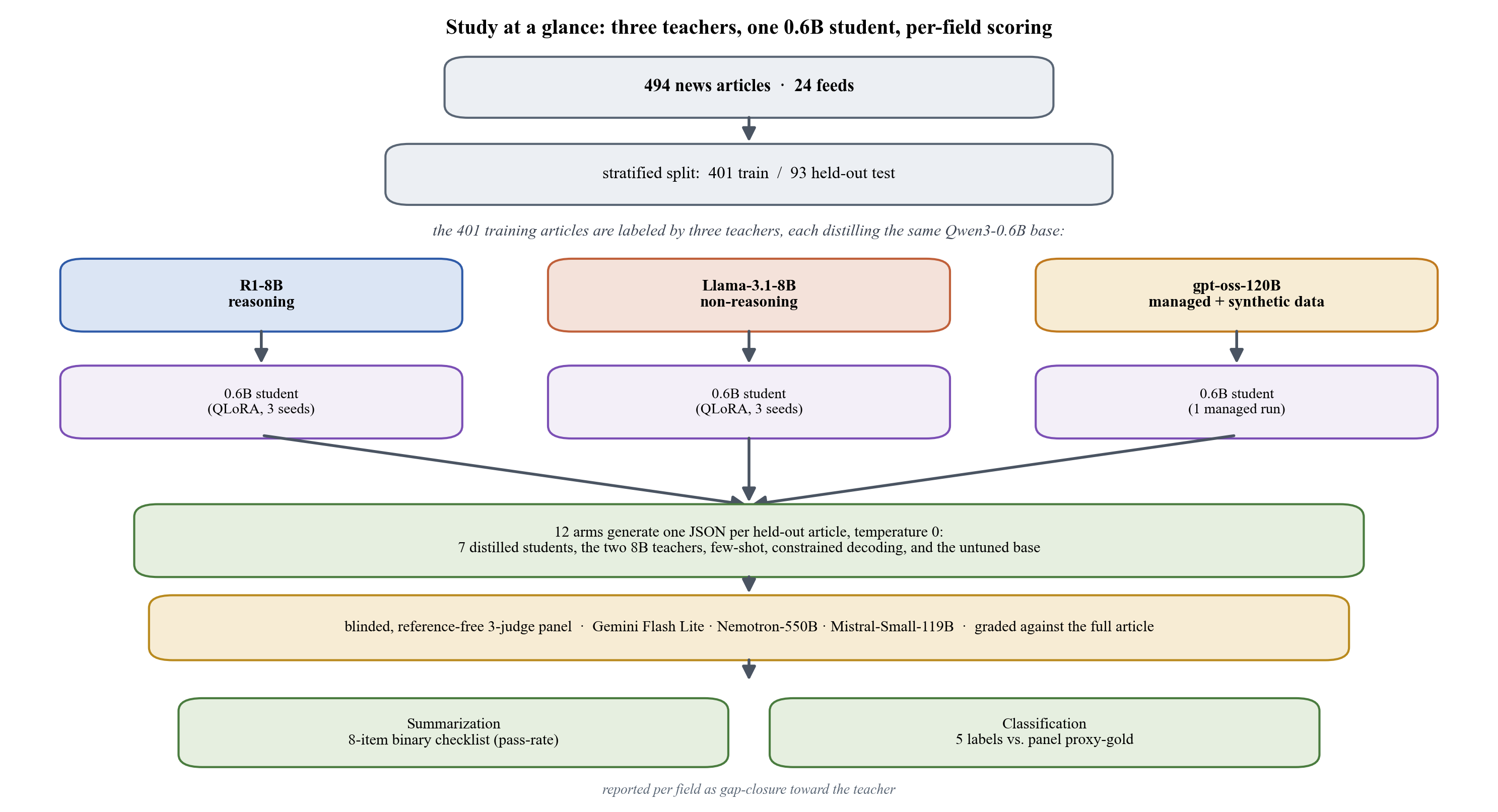}
\caption{The experimental pipeline. 494 labeled articles are split 401/93; three teachers, a reasoning 8B, a same-size non-reasoning 8B, and a larger managed pipeline with synthetic-data expansion, each distill the same Qwen3-0.6B base. Twelve arms (the seven distilled students, the two 8B teachers, two non-distillation controls, and the untuned base) generate on the held-out set and are scored on two sub-tasks by a blinded three-judge panel against the full article, reported as gap-closure toward the teacher.}
\label{fig:glance}
\end{figure*}

\textbf{Models.} The teacher is DeepSeek-R1 8B (\texttt{deepseek-r1:8b}); the student is Qwen3-0.6B. Both are served on-device at Q4\_K\_M quantization. A {\raise.17ex\hbox{$\scriptstyle\sim$}}13$\times$ parameter gap and a difference in kind: DeepSeek-R1 emits chain-of-thought before answering, whereas the student is trained on the teacher's final outputs only, compiling task knowledge into weights and skipping runtime deliberation. This choice is deployment-realistic; whether the teacher's reasoning nature, as opposed to its 8B scale, is what the student actually inherits is settled separately by a same-size non-reasoning-teacher control (Section~\ref{sec:reasoning}).

\textbf{Data and split.} 500 articles were sampled fairly across all feeds; the teacher generated a gold JSON for each; the six whose JSON was malformed were dropped, leaving 494 labeled articles. A stratified-random split (test fraction 0.2, every feed represented) gives 401 train / 93 test. The test set is new articles from known feeds, matching the intended use, new items from a fixed set of feeds, not generalization to unseen sources, which is out of scope.

\textbf{Corpus characteristics.} The dataset, which we call RSS-News, shapes several of this study's central findings, so its properties are reported explicitly (Table~\ref{tab:corpus}). The 494 articles span 24 feeds; sampling was balanced (27 articles for each of the 16 largest feeds, tapering to 2--3 for the smallest), so no single source dominates. The subject mix is technology- and AI-leaning, AI research and news, software and web development, cybersecurity, alongside general-interest long-form (\emph{The Atlantic}, \emph{Longreads}, \emph{The Marginalian}, \emph{Aeon}). Two properties bear directly on the results. First, article length is bimodal and capped: full text is stored up to {\raise.17ex\hbox{$\scriptstyle\sim$}}4\,096 characters (median 4\,040), but 22 of the 93 test articles ({\raise.17ex\hbox{$\scriptstyle\sim$}}24\%) arrive as short RSS excerpts of $\leq$1\,200 characters, the thin-source regime where the faithfulness regression concentrates (Section~\ref{sec:faithful}). Second, every categorical vocabulary is imbalanced in the teacher's own labels (Table~\ref{tab:corpus}): \emph{depth} is 71\% \emph{standard} with \emph{deep\_dive} essentially absent (0.8\%), \emph{urgency} is dominated by \emph{developing}/\emph{evergreen} with \emph{breaking} at 1.6\%, and \emph{tone} is 68\% \emph{analytical}. These skews, not raw article count, bound what the categorical fields can learn, and are the reference for the majority-class baselines in Section~\ref{sec:classification}. Fewer than 1\% of teacher labels fell outside the closed vocabularies (e.g., a stray \emph{humourous} \emph{frame}); these off-vocabulary values were retained as-is rather than corrected, and are a minor source of proxy-gold noise. Topics are genuinely open-set: 1\,359 distinct tags over 494 articles, a mean of 3.4 per article. Teacher summaries are a median of three sentences (431 characters).

\begin{table}[t]
\centering
\caption{RSS-News Corpus Characteristics (494 Labeled Articles, 24 Feeds)}
\label{tab:corpus}
\small
\setlength{\tabcolsep}{3pt}\begin{tabular}{@{}l p{5.2cm}@{}}
\toprule
Property & Value \\
\midrule
Articles (train / test) & 494 (401 / 93), stratified by feed \\
Feeds & 24; balanced ($\leq$27/feed), range 2--27 \\
Article length (chars) & median 4\,040; capped {\raise.17ex\hbox{$\scriptstyle\sim$}}4\,096; min 197 \\
Short excerpts ($\leq$1\,200) & 22/93 test ({\raise.17ex\hbox{$\scriptstyle\sim$}}24\%) \\
\emph{sentiment} & neutral 46 / positive 44 / negative 9 \\
\emph{urgency} & developing 59 / evergreen 39 / breaking 2 \\
\emph{frame} & analytical 55 / conflict 24 / human 15 / econ.\ 5 \\
\emph{tone} & analytical 68 / optimistic 17 / opinion 8 / alarm 7 \\
\emph{depth} & standard 71 / brief 28 / deep\_dive 1 \\
Topics (open-set) & 1\,359 distinct; mean 3.4/article \\
Summary (teacher) & median 3 sentences / 431 chars \\
\bottomrule
\end{tabular}
\end{table}

\textbf{Training.} A QLoRA fine-tune of Qwen3-0.6B directly on the 401 teacher outputs (LoRA rank 32, response-only loss masking, thinking disabled), on a free cloud T4 GPU. Three seeds (42/123/7) are trained so that seed variance is characterized rather than assumed away. Models are exported to Q4\_K\_M and served locally.

\textbf{Arms.} All arms generate at temperature 0 and see the full article. The twelve arms are the two 8B teachers, reasoning (\texttt{deepseek-r1:8b}) and same-size non-reasoning (\texttt{llama-3.1-8b-instruct}), each graded on the test set as an arm; the untuned base Qwen3-0.6B; base + few-shot prompting (2--3 in-context examples, controlling for prompting); base + constrained JSON decoding (controlling for formatting); the reasoning-teacher student at three seeds; the non-reasoning-teacher student at three seeds (identical recipe and seeds; Section~\ref{sec:reasoning}); and a managed-pipeline student (Section~\ref{sec:platform}) trained from the same 401-item upload through a commercial service with a larger teacher (\texttt{gpt-oss-120b}) and synthetic-data expansion, a single run. Distillation is credited only where it beats \emph{both} non-distillation controls on something real.

\section{Evaluation Method}\label{sec:eval}
\subsection{Metrics per Sub-Task}
\emph{Structure} is schema validity: parses as JSON with all seven fields, computed deterministically. \emph{Classification} is per-field accuracy plus macro-average against a panel-consensus proxy-gold (Section~\ref{sec:judges}; the majority-class baseline of Section~\ref{sec:classification} is the reference line). \emph{Summarization} is an eight-item binary checklist (pass-rate), the primary metric, chosen because a 0--5 rubric saturates; the checks are \emph{faithful}, \emph{thesis}, \emph{takeaway}, \emph{length}, \emph{opening}, \emph{teacher-lens}, \emph{tech-lens}, and \emph{tone}, with a separately reported \emph{topics\_cover} check (full wording in Appendix~\ref{app:checklist}). \emph{Efficiency} is latency, throughput, and memory, measured per-arm. The checklist and primary comparison were set after a 12-article pilot, before scoring; two checks were dropped at the pilot for lack of headroom.

\subsection{Judge Panel and Grading}\label{sec:judges}
Grading is reference-free, against the full article, never against the teacher's answer (which would induce teacher-mimicry bias); blinded, with shuffled anonymous arm labels; and at temperature 0. The panel deliberately excludes the teacher's and student's families, since evaluators favor their own and same-family generations \cite{panickssery2024, wataoka2024}. Results come from a three-judge panel, Gemini Flash Lite, Nemotron-550B, and Mistral-Small-119B, across three distinct families (Google, NVIDIA, Mistral), $N=93$; a panel of diverse models is more reliable than any single judge \cite{verga2024poll}. Groq-hosted judges were evaluated during panel selection but excluded for free-tier instability. Three judges give a true majority on every checklist item and classification label, so both the summary pass-rates and the classification proxy-gold are majority votes with no tie-breaking heuristic. Per-judge robustness is still reported (Section~\ref{sec:robust}): a finding is trusted when it survives under each judge alone. Grading uses the full article throughout: partial-context grading systematically understates faithfulness and manufactures an apparent regression, quantified in Section~\ref{sec:robust}.

\subsection{Grader Validation Without Humans}\label{sec:negcontrol}
At this scale, human adjudication is the bottleneck, so the grader is validated by construction: a negative control grades a sample of summaries against a \emph{mismatched} article. A grader that cannot tell a real summary from a mismatched one is not measuring anything. The result is 0\% faithful on mismatched articles ($n=30$): the grader demonstrably discriminates the gross case. It does \emph{not} confirm detection of subtle, in-domain fabrication, a plausible invented detail inside an otherwise-correct summary, which would require injected-fabrication tests or a human-labeled set (Section~\ref{sec:future}). The faithfulness numbers should be read with that boundary in mind.

\subsection{Statistics}\label{sec:stats}
Metrics are reported as mean with 95\% confidence interval (bootstrap over the 93 test items; headline comparisons use a paired bootstrap over per-article scores, 20\,000 resamples, cancelling per-article difficulty). Tuned metrics are the mean across three seeds. The primary comparison is tuned versus base + constrained on checklist pass-rate. Secondary comparisons (tuned versus base, tuned versus few-shot, and the per-field classification comparisons) are numerous, so individual secondary wins not also confirmed under each judge alone (Section~\ref{sec:robust}) are treated as suggestive. The per-article bootstrap captures article-level variance around the three-seed mean; it does not fully propagate training-seed variance, which Section~\ref{sec:seeds} shows is first-order on the subjective fields, there, the seed range, not the bootstrap interval, is the representative uncertainty.

\section{Results}\label{sec:results}
Results are organized by sub-task. A per-axis overview (Section~\ref{sec:gapclosure}) opens; summarization (Section~\ref{sec:summ}) then tells one continuous arc, the checklist win, aggregate faithfulness, and the thin-source length effect that is the same phenomenon seen twice; classification and its seed variance follow (Section~\ref{sec:classification}); Section~\ref{sec:reasoning} tells the capability split across all three teachers start to finish; and structure, topics, and efficiency close (Section~\ref{sec:efficiency}).

\subsection{Gap-Closure Toward the Teacher}\label{sec:gapclosure}
\begin{figure}[t]\centering
\includegraphics[width=\columnwidth]{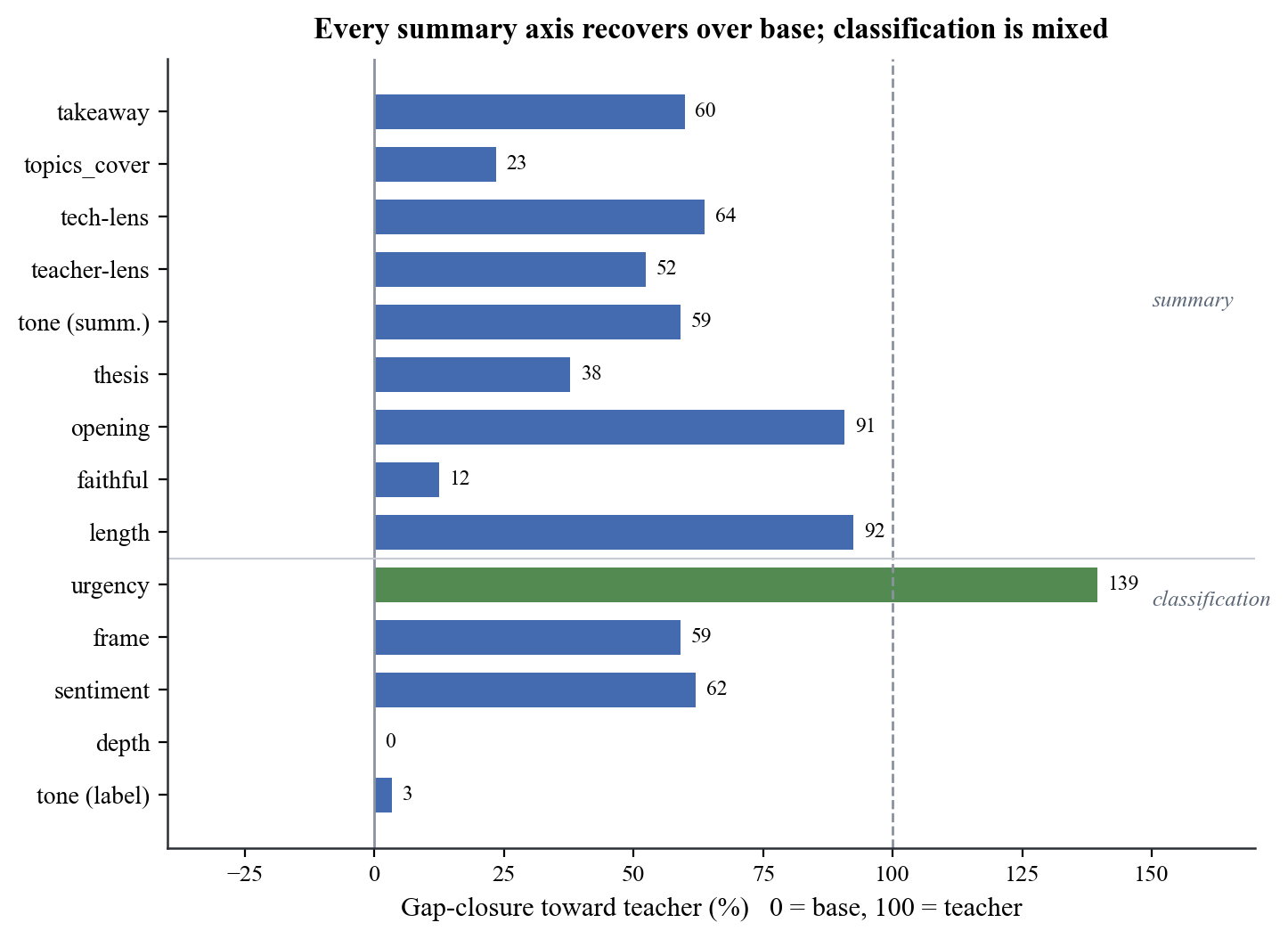}
\caption{Per-axis gap-closure toward the teacher (0 = untuned base, 100 = teacher). Under the canonical three-judge grading every summary axis is recovered over base (none falls below it); classification is more mixed, with \emph{urgency} crossing the teacher line only because the proxy-gold is skewed and \emph{depth} flat. The per-axis pattern, not any single aggregate, is the primary result.}
\label{fig:gap}
\end{figure}

Because the goal was teacher quality at student speed, the natural scale is how far the student traveled from the untuned base (0\%) toward the teacher (100\%) on each axis: $(\text{tuned}-\text{base})/(\text{teacher}-\text{base})$. Fig.~\ref{fig:gap} shows the shape of recovery, which matters more than any single number: distillation recovered every summary axis over base, from 12\% of the gap on \emph{faithful} to 92\% on \emph{length}, with no axis left below base. The one residual concern is short-source faithfulness, a subgroup effect the per-axis averages cannot show (Section~\ref{sec:summ}). On classification the picture is more mixed (Section~\ref{sec:classification}). Gap-closure is a ratio and becomes unstable when teacher and base are close, so absolute deltas are reported alongside every gap-closure figure and axes where teacher $\approx$ base are flagged; because the student can exceed 100\% on some axes (\emph{urgency}, an artifact of a skewed gold), the teacher is a reference point, not a ceiling.

\subsection{Summarization}\label{sec:summ}
Checklist pass-rates (Table~\ref{tab:arms}) show a decisive, significant primary win: tuned versus base + constrained is 75.1 versus 58.2, $+16.8$ points, paired bootstrap $[11.4, 22.5]$, $p<0.001$, holding for each of the three seeds independently. Distillation also beats the prompting control, 75.1 versus 70.2, $+4.9$ $[1.6, 8.2]$, $p=0.003$. The gain therefore survives both controls a reviewer would demand: it is not reducible to schema formatting or to a few in-context examples. Overall the tuned student closes 58\% of the base-to-teacher summary gap.

\begin{table*}[t]
\centering
\caption{All Twelve Arms on the 93-Item Test Set (Canonical Three-Judge Grading). Latency is per-article p50; the reasoning teacher runs on-device at {\raise.17ex\hbox{$\scriptstyle\sim$}}39 s.}
\label{tab:arms}
\small
\begin{tabular}{@{}lrrrrr@{}}
\toprule
Arm & Schema \% & Checklist \% & Faithful \% & Macro \% & p50 (ms) \\
\midrule
R1-8B teacher (reasoning) & 100 & 88.6 & 96 & 67.7 & {\raise.17ex\hbox{$\scriptstyle\sim$}}39{,}200 \\
Llama-8B teacher (non-reas.) & 98.9 & 68.8 & 96 & 67.8 & 4{,}713 \\
\midrule
R1 student, seed 42 & 100 & 74.6 & 77 & 64.5 & 748 \\
R1 student, seed 123 & 100 & 76.1 & 69 & 46.9 & 806 \\
R1 student, seed 7 & 100 & 74.5 & 73 & 54.6 & 768 \\
Llama student, seed 42 & 100 & 55.5 & 81 & 61.1 & 743 \\
Llama student, seed 123 & 100 & 58.6 & 77 & 53.5 & 761 \\
Llama student, seed 7 & 100 & 56.6 & 75 & 58.9 & 738 \\
Managed (gpt-oss-120B) & 98.9 & 71.0 & 66 & 66.3 & 929 \\
\midrule
Base (zero-shot) & 92.5 & 56.3 & 70 & 47.9 & 845 \\
Base + few-shot & 97.8 & 70.2 & 83 & 58.5 & 856 \\
Base + constrained & 100 & 58.2 & 73 & 48.2 & 824 \\
\bottomrule
\end{tabular}
\end{table*}

The per-check breakdown (Table~\ref{tab:percheck}) shows the gain is broad and, under the canonical grading, carries no aggregate regression: every check improves over base, from specificity (\emph{takeaway} 38 to 69) and the persona lenses through faithfulness ($+3$) and length ($+26$). The product-critical faithfulness check sits slightly \emph{above} base in aggregate and within the instrument's noise. The one real concern is invisible in these article-averaged numbers because it is a subgroup effect, faithfulness on short, thin-source articles, isolated next. So the honest headline is not ``the student regressed on faithfulness'' but ``the student improved everywhere the aggregate can see, and the residual risk is confined to a length-defined subgroup.''

\begin{table}[t]
\centering
\caption{Per-Check Summary Pass-Rates and Absolute Deltas}
\label{tab:percheck}
\small
\footnotesize\setlength{\tabcolsep}{4.5pt}\begin{tabular}{@{}lccccr@{}}
\toprule
Check & Base & Few & Tuned $\mu$ & Teach. & $\Delta$ base \\
\midrule
faithful & 69.9 & 82.8 & 73.1 & 95.7 & $+3.2$ \\
takeaway & 37.6 & 36.6 & 68.5 & 89.2 & $+30.9$ \\
teacher-lens & 21.5 & 37.6 & 50.2 & 76.3 & $+28.7$ \\
tech-lens & 33.3 & 43.0 & 48.4 & 57.0 & $+15.1$ \\
tone & 57.0 & 80.6 & 79.2 & 94.6 & $+22.2$ \\
opening & 80.6 & 97.8 & 98.2 & 100.0 & $+17.6$ \\
thesis & 80.6 & 87.1 & 87.1 & 97.8 & $+6.5$ \\
length & 69.9 & 95.7 & 95.7 & 97.8 & $+25.8$ \\
topics\_cover & 78.5 & 81.7 & 82.8 & 96.8 & $+4.3$ \\
\bottomrule
\end{tabular}
\end{table}

\textbf{Aggregate faithfulness holds; the risk is a thin-source subgroup.}\label{sec:faithful}
Under the canonical full-source grading, aggregate faithfulness is base 69.9, tuned 73.1, teacher 95.7: the tuned student sits at or slightly above base, so the aggregate ``faithfulness regression'' that coarser gradings suggest does not survive here. What is robust is a subgroup effect tied to article length. On long articles ($>1200$ chars, $n=71$) the tuned student is above base (78.6 versus 71.9); on short articles ($\leq1200$, $n=22$) it drops below (55.0 versus 64.0). The mechanism is thin source: given little article to ground in, the reasoning-teacher student fills the summary with specific-sounding but unsupported claims (the thin-source example below); on rich source material it stays grounded. This is a property of the teacher's lineage, not of small models in general: the same-size non-reasoning teacher's students stay faithful on exactly these short articles (74.2, Section~\ref{sec:reasoning}). The short-source magnitude is grading-sensitive (Section~\ref{sec:robust}) and rests on 22 articles, so we read it as a direction, not a fixed number. For a user-facing pipeline that cannot tolerate fabrication, even a localized dip is worth routing around (Section~\ref{sec:assignment}).

\textbf{Article length: thin sources are hard for the student.}\label{sec:length}
The faithfulness dip on short articles is one facet of a broader effect: the 0.6B student degrades on thin sources across nearly every summary axis, while the teacher barely moves. Splitting the 93 test articles at 1200 characters (Table~\ref{tab:length}), the tuned student's overall checklist falls from 78 on long articles to 66 on short ($-12$), whereas the teacher falls only 89 to 86 ($-3$) and the untuned base is flat (57 to 55). The drop is not confined to faithfulness: the teacher-lens ($-34$), faithfulness ($-24$), tone ($-15$), and takeaway ($-12$) checks all fall sharply on short inputs, while the format checks (length, opening) and thesis capture hold. The mechanism is direct: given little source to work from, a small student cannot ground, expand, or adopt the teacher's voice, and substitutes fabrication and generic phrasing. A larger teacher does not need the material. Article length, not model choice alone, gates deployment: the routing table (Section~\ref{sec:routing}) sends thin sources to a larger engine precisely because that is where the student's recovery collapses.

\begin{table}[t]
\centering
\caption{Tuned Student: Summary Checks by Article Length (short $\leq$1200 chars vs.\ long; $N=93$)}
\label{tab:length}
\small
\begin{tabular}{@{}lccc@{}}
\toprule
Check & Short ($n$=22) & Long ($n$=71) & Drop \\
\midrule
teacher-lens & 24 & 58 & $-34$ \\
faithful & 55 & 79 & $-24$ \\
tone & 68 & 83 & $-15$ \\
takeaway & 59 & 71 & $-12$ \\
tech-lens & 42 & 50 & $-8$ \\
opening & 97 & 99 & $-2$ \\
length & 94 & 96 & $-2$ \\
thesis & 91 & 86 & $+5$ \\
\midrule
\emph{overall (tuned)} & \emph{66} & \emph{78} & \emph{$-12$} \\
\emph{overall (teacher)} & \emph{86} & \emph{89} & \emph{$-3$} \\
\bottomrule
\end{tabular}
\end{table}

\textbf{Qualitative: a clean summary win.} The untuned base's characteristic failure is to spill the label values into the summary prose (``\ldots the sentiment is positive, urgency developing, frame human\_interest\ldots'') instead of writing a summary. The distilled student writes a faithful three-sentence summary that states the article's thesis and independently corrects an \emph{urgency} label the base got wrong. This is the modal difference behind the $+18.8$-point checklist gain over base.

\textbf{Qualitative: the thin-source failure.} On a short article the distilled student invents a framing absent from the source, a subject's presidency cast as an ``Obama-style conflict'', while the teacher stays grounded. This one instance is the short-source fabrication above, and the reason the routing table falls back to a larger engine on thin sources.

\subsection{Classification}\label{sec:classification}
Per-field accuracy against the panel proxy-gold (Table~\ref{tab:classification}) is read against a majority-class baseline, since several vocabularies are imbalanced and accuracy below that line is worse than a constant guess. \emph{Urgency} looks like a win, the tuned student reaches 72.4\%, above the teacher's 67.7, but the proxy-gold is heavily evergreen-skewed (majority baseline {\raise.17ex\hbox{$\scriptstyle\sim$}}74\%) and the student predicts evergreen on the large majority of items, so its urgency accuracy mostly reflects alignment with the skewed gold rather than learned discrimination; no arm reliably detects the rare \emph{breaking} class ($n=6$). \emph{Frame} shows a milder version of the same pattern. \emph{Depth} is a caution twice over: at 42.7 the tuned student sits below the {\raise.17ex\hbox{$\scriptstyle\sim$}}56\% of always guessing ``standard,'' and only the managed pipeline (\emph{Plat.}\ column, 59.8) clears the line; and \emph{depth} is also the least trustworthy gold, with inter-judge agreement of only 0.38 (Table~\ref{tab:classification}), so the ``below majority baseline'' verdict is a weak signal against a weak reference. That agreement fell from earlier two-judge grading because a third independent judge was added, not because the arms regressed. \emph{Tone} as a label barely moves under distillation (34.1 versus base 32.6) while few-shot reaches 74.7, a field cued far better in context than baked into weights. On the macro average, fine-tuning does not beat prompting (55.3 versus 58.5): on the aggregate categorical task the distillation advantage is confined to \emph{urgency} and \emph{frame}, while the managed pipeline (66.3) is the only small-model arm near the teachers. This is why the deployment recommendation is per-field (Section~\ref{sec:assignment}).

\begin{table}[t]
\centering
\caption{Per-Field Classification Accuracy vs.\ Panel Proxy-Gold (\emph{Plat.}\ = managed 120B pipeline)}
\label{tab:classification}
\small
\footnotesize\setlength{\tabcolsep}{3pt}\begin{tabular}{@{}lccccccc@{}}
\toprule
Field & Agr. & Maj. & Base & Few & Tuned $\mu$ & Teach. & Plat. \\
\midrule
urgency & 0.56 & {\raise.17ex\hbox{$\scriptstyle\sim$}}74 & 55.8 & 50.5 & 72.4 & 67.7 & 77.2 \\
frame & 0.70 &, & 51.2 & 39.6 & 58.4 & 63.4 & 65.2 \\
sentiment & 0.77 &, & 48.8 & 67.0 & 69.2 & 81.7 & 71.7 \\
depth & 0.38 & {\raise.17ex\hbox{$\scriptstyle\sim$}}56 & 51.2 & 60.4 & 42.7 & 48.4 & 59.8 \\
tone & 0.68 &, & 32.6 & 74.7 & 34.1 & 77.4 & 57.6 \\
macro &, &, & 47.9 & 58.5 & 55.3 & 67.7 & 66.3 \\
\bottomrule
\end{tabular}
\end{table}

\textbf{Seed variance.}\label{sec:seeds}
The three seeds diverge sharply on the hardest axis: \emph{tone}-label accuracy ranges 11.8\% to 58.1\% across seeds, and classification macro from 46.9 to 64.5. One seed (123) drives much of that range, posting macro 46.9 and \emph{tone} 11.8 against the other seeds' 58--65 and 32--58; it is not discarded, because that fragility is itself the finding. A single-seed study could have reported any point in that range as ``the'' result; at 0.6B, seed choice is first-order on the subjective fields, not a rounding error, which is why tuned values are three-seed means. Seed agreement is also a near-free confidence signal (Table~\ref{tab:seeds}): on four of the five fields, unanimous predictions are markedly more accurate than split ones, by up to 27 points on \emph{sentiment} and \emph{urgency}, and \emph{tone}, the field distillation fails, is where the seeds almost never agree (unanimous on 12 of 93). The exception is \emph{depth}, where unanimity does not help: the seeds agree confidently on a wrong majority label, consistent with \emph{depth} sitting below its majority-class baseline. Since the three seeds together still run in {\raise.17ex\hbox{$\scriptstyle\sim$}}2.4 s/article, seed-ensemble agreement is a per-item confidence flag at nearly no cost: trust unanimous labels, route split items to a fallback (Section~\ref{sec:assignment}).

\begin{table}[t]
\centering
\caption{Seed Unanimity Predicts Label Correctness}
\label{tab:seeds}
\small
\begin{tabular}{@{}lcccc@{}}
\toprule
Field & Unanimous (3/3) & Acc. & Split & Acc. \\
\midrule
urgency & 83 & 75.9 & 10 & 50.0 \\
sentiment & 67 & 77.6 & 26 & 50.0 \\
frame & 57 & 66.7 & 36 & 61.1 \\
depth & 36 & 44.4 & 57 & 52.6 \\
tone & 12 & 58.3 & 81 & 43.2 \\
\bottomrule
\end{tabular}
\end{table}

\textbf{Qualitative: prompting beats fine-tuning on tone.} On \emph{tone} the distilled student collapses onto one or two labels (34\% accuracy), while the base model given 2--3 in-context examples recovers the distinction (75\%). The label is cued far better by a few demonstrations than by weights compiled from the teacher's outputs, the concrete reason \emph{tone} is routed to few-shot in the shipped configuration (Section~\ref{sec:design}).

\subsection{What Each Teacher Transfers}\label{sec:reasoning}
\begin{figure}[t]\centering
\includegraphics[width=\columnwidth]{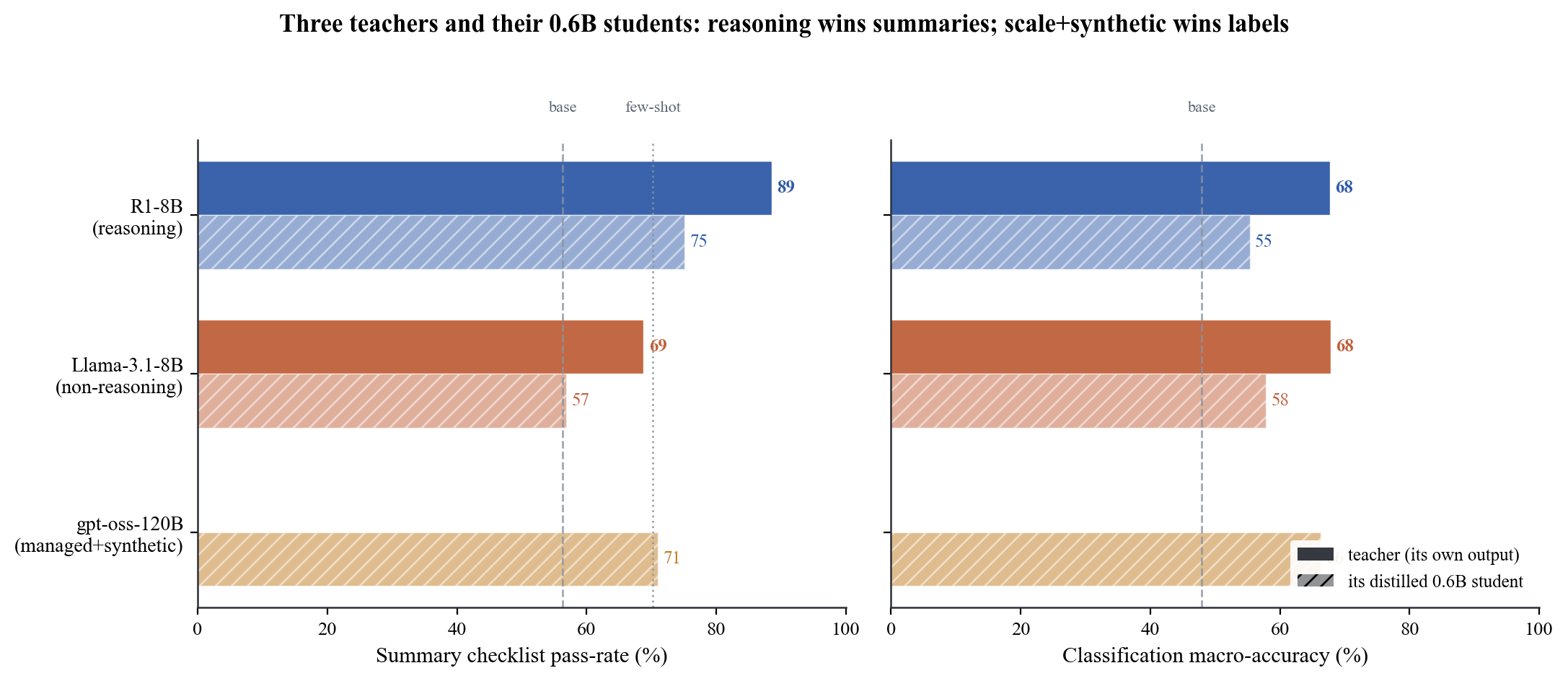}
\caption{Three teachers and their 0.6B students under one rubric (full 12-arm re-grade, $N=93$; students are three-seed means). Solid bars are a teacher's own output, hatched bars its distilled student. Left, summaries: the reasoning teacher (R1-8B) and its students dominate; the same-size non-reasoning teacher (Llama-3.1-8B) and its students sit at the untuned-base line; the managed 120B pipeline falls between. Right, classification: teachers and students are level and the managed pipeline leads. Reasoning transfers writing quality; scale with synthetic expansion transfers label diversity.}
\label{fig:reasoning}
\end{figure}

Three teachers distil the same Qwen3-0.6B base by the identical QLoRA recipe, seeds, and split, so that what each student inherits is attributable to the teacher and not the procedure. The reasoning teacher is DeepSeek-R1-Distill-Llama-8B; the same-size non-reasoning teacher is Llama-3.1-8B-Instruct, which re-labeled the 401 training articles under the identical prompt at temperature 0, yielding usable JSON on 396; the third is a managed pipeline running a larger model, \texttt{gpt-oss-120b}, with proprietary synthetic-data expansion. The two 8B teachers differ only in reasoning nature and isolate it; the managed pipeline additionally varies scale and adds synthetic data, so it is a systems comparison, not a clean control. Each 8B teacher's own outputs on the test set are graded as an arm, so a teacher sits beside its students under one rubric (Fig.~\ref{fig:reasoning}); all twelve arms were re-graded together and prior aggregates reproduce within {\raise.17ex\hbox{$\scriptstyle\sim$}}2 points.

\textbf{Reasoning is what the summary gain depends on.} The reasoning teacher out-writes the non-reasoning teacher on its own summaries, 88.6 versus 68.8 on the checklist, and that {\raise.17ex\hbox{$\scriptstyle\sim$}}20-point teacher-level gap propagates almost intact to the students: 75.1 versus 56.9, a paired $+18.1$ points, bootstrap $[14.4, 22.0]$, $p<0.001$, holding across all three seed pairs ($+19.1$, $+17.5$, $+17.9$). The comparison to no distillation makes it sharp: the reasoning student beats the untuned base by $+18.7$ $[13.1, 24.4]$, whereas the non-reasoning student does not beat base at all ($+0.6$ $[-5.2, 6.4]$, n.s.). A same-size non-reasoning teacher bought essentially nothing on summary quality; the student inherits its teacher's ceiling, and only the reasoning teacher's ceiling is high enough to move the result. That a stronger \emph{kind} of teacher rather than a stronger-scoring one is what transfers echoes reports that teacher task-fit, not benchmark score, governs distillation gains \cite{chen2025factors, ramesh2025}.

\textbf{The advantage is specificity, not faithfulness.} Two qualifications keep the claim precise. The gain is not an artifact of a rubric that rewards the teacher's persona: on the three article-grounded, persona-free checks, \emph{faithful}, \emph{thesis}, \emph{takeaway}, the reasoning student still leads 76.2 versus 57.0, $+19.2$ $[14.2, 24.4]$, $p<0.001$, driven by \emph{takeaway} specificity (68.5 versus 16.5), the reasoning teacher supplying concrete exemplars where the non-reasoning teacher supplied vague ones. But the advantage stops short of faithfulness: on the \emph{faithful} check alone the non-reasoning student is marginally higher, 77.8 versus 73.1 ($\Delta-4.7$ $[-11.8, 2.5]$, $p=0.19$, not significant), matching evidence that reasoning-oriented training can raise rather than lower hallucination \cite{aggarwal2025, yao2025}. And on closed-set labeling the two 8B teachers themselves tie (macro 67.8 versus 67.7), as do their students (57.8 versus 55.3): reasoning lifts the open-ended sub-task and leaves classification untouched. The reasoning teacher earns its place in the title on summaries, and there alone.

\textbf{The third teacher: scale and synthetic expansion.}\label{sec:platform}
The managed pipeline uploads the same 401 articles to a commercial service that runs a larger teacher (\texttt{gpt-oss-120b}), expands them synthetically, and fine-tunes the same base in one hands-off run. Scale, data recipe, and run count all differ from the two 8B teachers at once. Its student trails the reasoning student on summaries (71.0 versus 75.1) yet posts the best small-model classification macro, 66.3, against 55.3 and 57.8 for the two 8B-teacher students and close to the teachers themselves, and it alone among the students clears the \emph{depth} majority baseline (Table~\ref{tab:classification}). Scale with synthetic expansion transferred label diversity where the reasoning teacher transferred writing.

Across the three teachers (Fig.~\ref{fig:reasoning}) the capability split is clear, and no teacher wins everything. Two of the three patterns are strong: the reasoning teacher owns summary quality and the managed pipeline owns classification, each a significant effect. The third is weaker and stated as such. The same-size instruction teacher is the most grounded model measured (96\% faithful, 95\% even on short sources), and its students are the most grounded students, but the evidence is an ordering, not a significant gap: on aggregate faithfulness the instruction students lead the reasoning students only 77.8 versus 73.1, a difference the paired bootstrap does not resolve ($p=0.19$, above). The separation is clear only on the 22 short articles, where the instruction students hold 74 faithful while the reasoning students fall to 55 and the managed student collapses to 36, the worst in the study. Because those short-source magnitudes are grading-sensitive (Section~\ref{sec:robust}) and rest on 22 items, we report thin-source grounding as a consistent direction that tracks the teacher's lineage, not as a measured transfer coefficient. It nonetheless has an operational consequence: the most grounded thin-source engine is the instruction teacher, not the reasoning one, which the routing analysis takes up (Section~\ref{sec:routing}).

\subsection{Structure, Topics, and Efficiency}\label{sec:efficiency}
\begin{figure}[t]\centering
\includegraphics[width=\columnwidth]{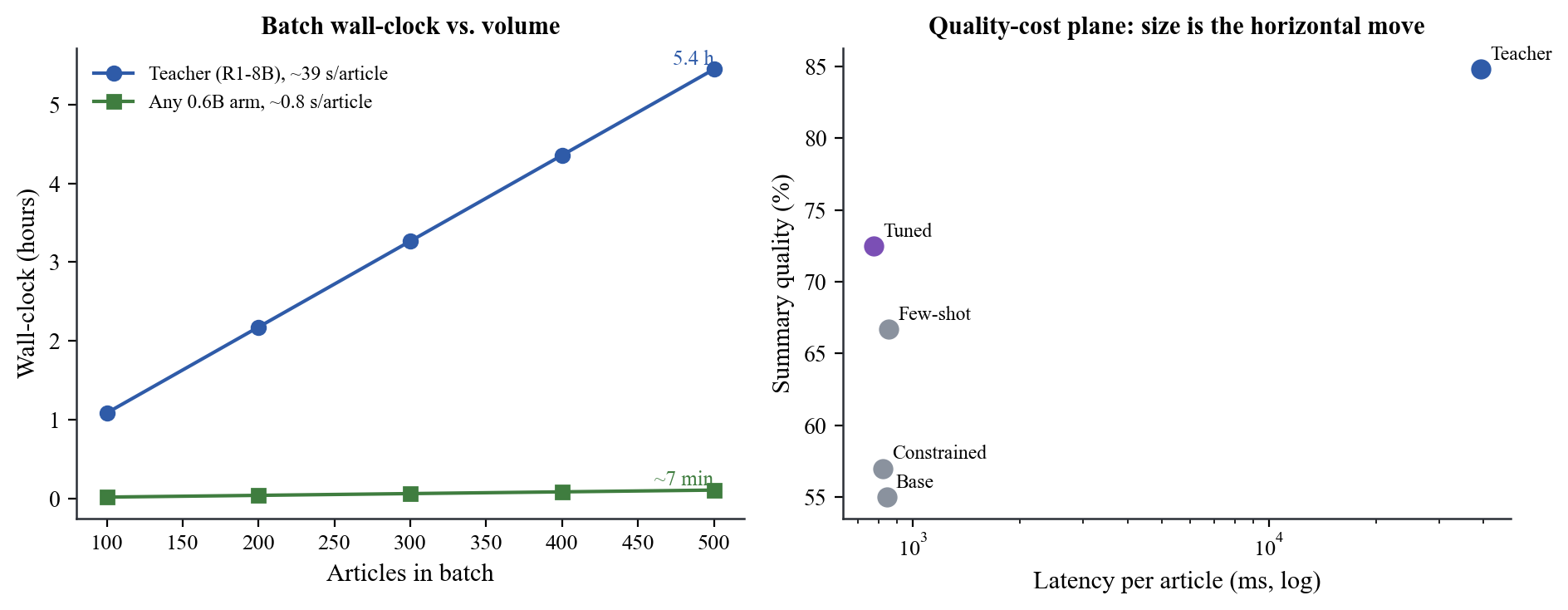}
\caption{Left: batch wall-clock versus volume, the 5.4 h to {\raise.17ex\hbox{$\scriptstyle\sim$}}7 min collapse at 500 articles that motivated the work. Right: the quality-cost plane, every 0.6B arm sits at {\raise.17ex\hbox{$\scriptstyle\sim$}}0.8 s; the vertical axis is where distillation, prompting, and the teacher differ.}
\label{fig:efficiency}
\end{figure}

The tuned student produces valid seven-field JSON reliably, but constrained decoding reaches the same validity with zero training, so structure is a decoding flag, not a reason to distill. Topic coverage is tuned 82.8\% versus base 78.5\% versus teacher 96.8\%, a solid low-stakes gain. On efficiency (Table~\ref{tab:arms}, Fig.~\ref{fig:efficiency}), every 0.6B arm runs at {\raise.17ex\hbox{$\scriptstyle\sim$}}0.8 s/article; the tuned model is fastest by wall-clock, not because throughput is higher but because distillation taught it to write shorter outputs. The 5.4 h to {\raise.17ex\hbox{$\scriptstyle\sim$}}7 min batch collapse is real and slightly better for the distilled model than for any other student arm.

\section{Grading Validity}\label{sec:robust}
This section reports the two grading-setup effects the study depends on, the full-source requirement and the panel's per-judge stability, and leads with the one that transfers beyond this paper.

\emph{Reference-free faithfulness grading must use the full source.} This is the single most transferable methodological result of the study. Grading against only the first 1200 characters produces a large spurious faithfulness deficit for the tuned student that full-article grading removes, leaving only the genuine short-article effect (Section~\ref{sec:faithful}). Two checks confirm the artifact: the lead-only-to-full uplift appears for \emph{every} arm, a grading-setup effect, not a model regression, and on long articles the tuned student is at or above base, while short articles, where the two contexts coincide, are unaffected: a built-in control. The lesson generalizes: grade reference-free faithfulness against the full source, or a truncation artifact masquerades as a model regression larger than the real effect.

\emph{Per-judge robustness.} The three judges disagree on individual items, so magnitudes are panel-dependent, but every directional finding in Section~\ref{sec:results} holds under each judge alone: all three rank the tuned student above both non-distillation controls on the checklist, all three place the reasoning-teacher student far above the non-reasoning one, and all three rank \emph{tone}-label few-shot far above tuned. Only directional claims are made on judge-graded magnitudes; only human gold labels can settle the exact values (Section~\ref{sec:limits}). Notably, the judges disagree even on mechanically checkable predicates such as sentence count and opening phrase, confirming that rule-checkable checks are better computed by rule than judged; and any verbosity bias runs \emph{against} this study's conclusion, since the winning arm writes the shortest outputs.

\emph{The faithful check is composition-sensitive.} It is the most sensitive of the eight to prompt composition: adding or removing an arm reshuffles the blinded label order in every batched prompt, and the \emph{faithful} value then moves several points on some arms while the other checks stay put. This is why the short-source faithfulness gap is reported as a direction, not a fixed number.

\section{Per-Field Engine Assignment}\label{sec:assignment}
Judged against the goal of teacher-quality output at student speed: \emph{speed} is fully met ({\raise.17ex\hbox{$\scriptstyle\sim$}}0.8 s/article, 5.4 h to {\raise.17ex\hbox{$\scriptstyle\sim$}}7 min, on-device, at no cost); \emph{summary quality} is substantially met (58\% gap closure, significant wins over both controls), with one localized caveat, faithfulness on short-source articles; \emph{classification quality} is met unevenly (about a third of the gap closed, the macro does not beat few-shot, \emph{tone} barely moves, and \emph{depth} sits below its majority-class baseline). The measured result, and the transferable idea, is therefore not ``use the distilled model'' but a per-field routing map (Table~\ref{tab:assignment}). The map is a measurement deliverable; Section~\ref{sec:design} reduces it to the configuration that should ship.

\begin{table}[t]
\centering
\caption{Per-Field Best On-Device Engine}
\label{tab:assignment}
\small
\begin{tabular}{@{}p{1.9cm}p{2.1cm}p{2.8cm}@{}}
\toprule
Field / output & Best engine & Why \\
\midrule
JSON structure & constrained decoding & free; no training needed \\
urgency, frame & distilled student & best judge agreement; beats both controls \\
sentiment, topics & distilled, platform, or few-shot & roughly tied (67--72); platform edges sentiment; sub-second \\
depth & platform-distilled & only arm above the majority-class line \\
tone label & few-shot base & prompting far better here \\
summary, long/rich & distilled student & 58\% gap closed; faithful level with base \\
summary, short/thin & larger engine (R1 or Llama-8B) & fabrication concentrates here; R1 for writing, Llama-8B for faithfulness (Section~\ref{sec:routing}) \\
\bottomrule
\end{tabular}
\end{table}

The task pattern was chosen because it recurs: the batch collapse is the signature of any high-volume, schema-bound, latency-sensitive enrichment pipeline paying mid- or large-model latency per item, ticket triage (priority + category, the direct analogue of \emph{urgency} + \emph{frame}), log classification, document intake, moderation pre-filters, catalog extraction. For each, the recipe is the same: distill the small model for the structured, high-frequency labels; keep the large model for low-frequency, high-stakes free text; and measure per field rather than assuming the small model wins everything. One task family has been measured; the harness, not this routing table, is the artifact built to transfer.

\subsection{The Routing Table, Scored as a System}\label{sec:routing}
A recommendation assembled from per-field verdicts can and should be scored as a system. Because every arm's output for every test article was individually graded, any deterministic routing rule selects, for each article, an output that already carries grades, so the assembled system's metrics are exact arithmetic over existing per-item scores and the paired-bootstrap machinery applies unchanged. Four summary-routing variants are scored (Table~\ref{tab:routing}; labels always route per-field). The results are: (1) pure on-device fallbacks for short sources trade a little summary quality for a little faithfulness, Routers A and C nudge faithfulness up while costing 1--3 checklist points, so neither is a clean win; keep all-tuned, or pay a larger engine on short sources; (2) routing only the 22/93 short articles to a larger engine is the sound move for thin sources, and which engine is itself a capability choice. Falling back to the reasoning teacher (Router B) buys $+4.6$ checklist points over all-tuned at {\raise.17ex\hbox{$\scriptstyle\sim$}}82 minutes per 500 articles; falling back to the same-size instruction teacher instead (Router B$'$) gives the highest faithfulness of any configuration that keeps the bulk of traffic on-device (82.8) at {\raise.17ex\hbox{$\scriptstyle\sim$}}14 minutes, roughly six times faster, because that teacher is the most grounded model measured (Section~\ref{sec:reasoning}), but costs 3.4 checklist points because it writes plainer summaries. The thin-source fallback is therefore faithfulness-first (B$'$) or writing-quality-first (B), a direct instance of the capability split; (3) per-field classification routing, an oracle that picks the best of all six measured arms for each label post hoc (the hosted teachers included), reaches macro 76.1 against the teacher's own 67.7. This is an upper bound on what field-level routing could achieve, not a held-out result, and is labeled as such wherever it appears. These summary-routing differences ride on the 22-article short subgroup, and the routing rules were fixed after seeing the per-field results, so this is engineering validation, not a confirmatory finding.

\begin{table}[t]
\centering
\caption{Routing Configurations Scored as Assembled Systems}
\label{tab:routing}
\small
\setlength{\tabcolsep}{4pt}\begin{tabular}{@{}lcccc@{}}
\toprule
Configuration & Chk. & $\Delta$tuned & Faith. & Batch/500 \\
\midrule
All-tuned & 75.1 &, & 73.1 & {\raise.17ex\hbox{$\scriptstyle\sim$}}7 min \\
A: short$\rightarrow$few-shot & 74.0 & $-1.0$ & 75.3 & {\raise.17ex\hbox{$\scriptstyle\sim$}}7 min \\
C: short$\rightarrow$base & 72.4 & $-2.6$ & 75.3 & {\raise.17ex\hbox{$\scriptstyle\sim$}}7 min \\
B: short$\rightarrow$R1 teach. & 79.7 & $+4.6$ & 80.7 & {\raise.17ex\hbox{$\scriptstyle\sim$}}82 min \\
B$'$: short$\rightarrow$Llama & 71.6 & $-3.4$ & 82.8 & {\raise.17ex\hbox{$\scriptstyle\sim$}}14 min \\
All-teacher & 88.6 &, & 95.7 & 5.4 h \\
\bottomrule
\end{tabular}
\end{table}

\subsection{From Measurement to Solution Design}\label{sec:design}
The routing map answers a measurement question: which engine is best on each field, and by how much. A deployment answers a different question: which of those wins are worth operating. Measurement should be exhaustive; design should be ruthless. The design rule is the product of three measured quantities: route a field only where (accuracy gain from routing) $\times$ (cost of that field being wrong) exceeds the standing cost of one more engine to version, monitor, and re-evaluate. Applied here, the recommended configuration is two engines and one rule, the tuned student everywhere, a single few-shot pass for \emph{tone} (a $+50$-point gain at low failure cost), with the teacher fallback for thin-source summaries held as a deliberate, latency-priced option, not the full map. That captures most of the measured routing benefit at a fraction of the operating surface. Two costs this harness does not measure are stated plainly: cross-field coherence (nothing enforces agreement between a routed label and the summary), and maintenance surface (every routed engine is one more artifact to re-evaluate on any change). The distinction generalizes: the scientific value of a per-field evaluation is the complete map; the application value is the smallest routed subset that pays.

\section{Discussion and Methods Contribution}\label{sec:methods}
Beyond the routing table, the study's methodological contribution is the harness itself, enumerated in the contributions of Section~\ref{sec:intro} and detailed in Sections~\ref{sec:eval}--\ref{sec:robust}. The value is not any single ingredient, each of which is established in the judging literature (Section~\ref{sec:related}), but their assembly into a discipline that catches the ways a less controlled evaluation would overstate the method. That discipline surfaced as a chain of build decisions any team distilling for a production task must cross (Table~\ref{tab:decisions}); recording them is much of the exercise's transferable value.

\begin{table}[t]
\centering
\caption{The Decision Surface, as Practitioner Guidance}
\label{tab:decisions}
\small
\setlength{\tabcolsep}{4pt}\begin{tabular}{@{}p{2.0cm}p{5.05cm}@{}}
\toprule
Decision & What the study taught \\
\midrule
Teacher & Output you would ship is the bar; a same-size non-reasoning teacher does not match it on summaries, the reasoning nature is what transfers there (Section~\ref{sec:reasoning}) \\
Student size & Viable for style, structure, some labels, but seed variance is first-order; budget $\geq3$ seeds \\
Quantization & Measure at the quantization you will ship \\
Tuning recipe & The cheapest recipe moved summary quality decisively; evaluation rigor, not recipe cost, was the bottleneck \\
Training-set size & Enough for style and structure; class balance, not raw count, bound the subjective labels \\
Task framing & The single most consequential evaluation decision: real findings are invisible in the aggregate \\
Controls & Nearly free, and they changed the conclusion twice \\
Judges & Grade against the full source or a truncation artifact masquerades as a regression \\
Ship decision & The unit of measurement is the field; the unit of deployment is the minimal routing that pays \\
Build vs.\ buy & The managed pipeline trails on summaries but leads on classification; the recipes transfer different capabilities, choose per field \\
\bottomrule
\end{tabular}
\end{table}

\section{Limitations}\label{sec:limits}
The reasoning-teacher question is now addressed by a same-size non-reasoning-teacher control (Section~\ref{sec:reasoning}), which shows the summary gain is reasoning-specific; but that control varies the teacher's reasoning nature without holding constant every other difference between two distinct 8B models, their label distributions differ, and the non-reasoning teacher yielded 396 usable label sets against the reasoning teacher's 401, so it attributes the gain to the reasoning teacher as a whole, not to chain-of-thought as an isolated mechanism, and it rests on the same three-judge instrument as every other magnitude here. Labels are silver, not gold: classification accuracy is against a panel consensus, not human truth, so the numbers are judge priors, not ground truth. The three-judge panel gives a true majority but does not make magnitudes judge-invariant; a different panel would shift exact values, which is why per-judge robustness is reported and only directional claims are made on judge-graded magnitudes. Faithfulness is reference-free LLM-judged, not entailment-verified, and the negative control validates only the gross case, not subtle in-domain fabrication. The short-article finding rests on a small subgroup ($n=22$). Seed variance is first-order on subjective fields and is not fully propagated by the per-article bootstrap, and the platform arm is a single run without a seed-variance estimate of its own. The pipeline is third-order distillation, which may contribute to the ceiling, though student capacity is an equally plausible cause and we do not separate the two. Finally, a single task family and feed set is measured; no generalization to unseen sources is claimed.

\section{Future Work}\label{sec:future}
Ordered by expected value: (1) add a human-gold anchor on {\raise.17ex\hbox{$\scriptstyle\sim$}}50 items to convert directional magnitudes into settled numbers and calibrate the judges, the single change that would most raise the evidentiary standing of every quantitative claim here; (2) confirm the residual faithfulness gap with a formal entailment verifier and inject-fabrication tests; (3) turn the truncation observation into a measured curve; (4) fix the short-article fabrication mode via faithfulness-filtered data or a decoding-time guard; (5) recover \emph{tone} labeling with a class-balanced training set; (6) add a reasoning-trace transfer arm, testing whether supervising on the teacher's traces rather than its answers widens the reasoning-teacher advantage established here; and (7) identify which property of the reasoning teacher's labels, specificity, thesis selection, or structure, carries the summary gain isolated in Section~\ref{sec:reasoning}.

\section{Conclusion}\label{sec:conclusion}
A distilled 600M student runs the 500-article enrichment in {\raise.17ex\hbox{$\scriptstyle\sim$}}7 minutes instead of 5.4 hours, recovers most of the teacher's summary quality (58\% of the gap) and about a third of the classification gap, and beats both non-distillation controls on summary quality. The recovery is uneven in ways that matter: on classification the aggregate does not beat few-shot prompting, \emph{depth} is untrustworthy for every small-model arm, and on summaries the student matches or beats base on every axis with one residual risk, fabrication on short-source articles, a property of the reasoning teacher's lineage. The verdict is therefore not ``distillation works'' or ``distillation hallucinates'' but the two-layer answer developed in Section~\ref{sec:design}: an exhaustive per-field measurement map in which the winners differ, reduced to a deliberately smaller deployment, the tuned student everywhere, one few-shot pass for \emph{tone}, and a priced thin-source fallback. Two lessons generalize. Decompose before concluding, and check the instrument: scoring the sub-tasks apart made every real finding visible that an aggregate would hide, and grading against the full source rather than a truncated lead removed a spurious faithfulness ``regression'' and left only a genuine short-source effect, so in reference-free evaluation the measurement setup can be a larger effect than the model. The question the earlier version left open is now closed: a same-size non-reasoning teacher, distilled by the identical recipe, does not beat the untuned base on summaries, so it is the teacher's reasoning nature and not its scale that the student inherits, an inheritance of specificity and thesis-capture, not of faithfulness.

\appendices
\section{Summary Checklist and Judge Prompts}\label{app:checklist}
The checklist (eight binary checks plus topics), graded by the panel against the full article, is: \emph{faithful} (every factual claim supported by the article); \emph{thesis} (captures the central thesis); \emph{takeaway} (a concrete, specific takeaway); \emph{length} (3--4 sentences); \emph{opening} (does not begin with ``This/The article''); \emph{teacher-lens} (explains a concept accessibly); \emph{tech-lens} (addresses the technical angle); \emph{tone} (direct, contemplative, not alarmist); and \emph{topics\_cover} (the listed topics capture the main themes), reported separately. The checklist judge receives the full article and all arms' summaries under shuffled anonymous labels and returns strict JSON keyed by label. The classification judge assigns exactly one label per field from the closed vocabularies of Section~\ref{sec:task}; the panel vote is the proxy-gold. The negative control pairs each summary with a mismatched article; \emph{faithful} collapses to 0\% ($n=30$).

\section{Teacher-Comparison Protocol}\label{app:teachers}
Three teachers label the same 401 training articles under the identical system prompt at temperature 0, and the same Qwen3-0.6B base is distilled from each by the identical QLoRA recipe (rank 32, response-only loss, thinking disabled). The reasoning teacher is DeepSeek-R1-Distill-Llama-8B (\texttt{deepseek-r1:8b}); the same-size non-reasoning teacher is Llama-3.1-8B-Instruct, which produced usable JSON for 396 of the 401 articles against the reasoning teacher's 401; the managed pipeline uses \texttt{gpt-oss-120b} with proprietary synthetic-data expansion in a single hands-off run. The two 8B teachers differ only in reasoning nature, isolating that one variable; the managed pipeline additionally varies teacher scale and adds synthetic expansion, so it is a systems comparison rather than a clean control. Each 8B teacher's own outputs on the 93 test articles are graded as an arm under the same blinded checklist as its students; the managed pipeline's teacher is not graded directly, because its student was trained on synthetically expanded data rather than the teacher's raw labels, so a raw self-score would not be the signal the student actually saw. Adding any arm re-shuffles every blinded batched prompt, so all arms are re-graded together; reported aggregates reproduce within {\raise.17ex\hbox{$\scriptstyle\sim$}}2 points across successive re-grades.

\section{Reproducibility}\label{app:repro}
Every reported metric recomputes offline, with no API keys, from the released scorecard, the full-context multi-arm pass-rate file and the per-item per-check detail, via the significance, routing, and robustness scripts. The judge cache accelerates re-grading but is a speed cache keyed by prompt hash, not the canonical record; the released scorecard, not the cache, is the artifact the metrics reproduce from, and regenerating the grades themselves (rather than the metrics derived from them) requires the judge-panel clients and API keys. One boundary is stated plainly: the released grades are the three-judge panel majority per item; individual per-judge votes are retained only in the prompt-hashed cache, not in an item-attributable form, so the per-judge robustness of Section~\ref{sec:robust} was computed during the scored run and is not reconstructable from the released scorecard alone. Released artifacts include the test set with teacher gold, the 401/93 split, per-arm outputs with measured per-item latency (including the two 8B teachers' self-scored arms and both distilled-student families), the non-reasoning-teacher training set, the full-context multi-arm scorecard, the per-item per-check detail, and the analysis, teacher-contrast, routing, and neutral-subset scripts.

\section*{Acknowledgment}
The evaluation-harness code, the figure-generation scripts (which render measured data as plots; no figure content is AI-generated), and portions of the manuscript prose were produced with the assistance of AI coding and writing agents. All experimental design, the analysis plan, and the interpretation of results are the author's.


\begin{thebibliography}{99}
\bibitem{hinton2015} G.~Hinton, O.~Vinyals, and J.~Dean, ``Distilling the knowledge in a neural network,'' arXiv:1503.02531, 2015.
\bibitem{kimrush2016} Y.~Kim and A.~M.~Rush, ``Sequence-level knowledge distillation,'' in \emph{Proc. EMNLP}, 2016.
\bibitem{xusurvey2024} X.~Xu \emph{et al.}, ``A survey on knowledge distillation of large language models,'' arXiv:2402.13116, 2024.
\bibitem{zhu2023} X.~Zhu \emph{et al.}, ``A survey on model compression for large language models,'' arXiv:2308.07633, 2023.
\bibitem{yang2025} A.~Yang \emph{et al.}, ``Qwen3 technical report,'' arXiv:2505.09388, 2025.
\bibitem{liu2024mobilellm} Z.~Liu \emph{et al.}, ``MobileLLM: Optimizing sub-billion parameter language models for on-device use cases,'' arXiv:2402.14905, 2024.
\bibitem{rang2025} M.~Rang \emph{et al.}, ``Revealing the power of post-training for small language models via knowledge distillation,'' arXiv:2509.26497, 2025.
\bibitem{zhou2023universalner} W.~Zhou \emph{et al.}, ``UniversalNER: Targeted distillation from large language models for open named entity recognition,'' arXiv:2308.03279, 2023.
\bibitem{gu2023biomedical} Y.~Gu \emph{et al.}, ``Distilling large language models for biomedical knowledge extraction: A case study on adverse drug events,'' arXiv:2307.06439, 2023.
\bibitem{deepseek2025} DeepSeek-AI, ``DeepSeek-R1: Incentivizing reasoning capability in LLMs via reinforcement learning,'' arXiv:2501.12948, 2025.
\bibitem{zhang2025} C.~Zhang \emph{et al.}, ``100 days after DeepSeek-R1: A survey on replication studies and more directions for reasoning language models,'' arXiv:2505.00551, 2025.
\bibitem{li2025} Y.~Li \emph{et al.}, ``NaturalThoughts: Selecting and distilling reasoning traces for general reasoning tasks,'' arXiv:2507.01921, 2025.
\bibitem{chen2025factors} X.~Chen \emph{et al.}, ``Unveiling the key factors for distilling chain-of-thought reasoning,'' in \emph{Findings of the Assoc. Comput. Linguistics: ACL 2025}, 2025, arXiv:2502.18001.
\bibitem{ramesh2025} S.~K.~Ramesh \emph{et al.}, ``On the generalization vs fidelity paradox in knowledge distillation,'' in \emph{Findings of the Assoc. Comput. Linguistics: ACL 2025}, 2025, arXiv:2505.15442.
\bibitem{busbridge2025} D.~Busbridge \emph{et al.}, ``Distillation scaling laws,'' arXiv:2502.08606, 2025.
\bibitem{mirzadeh2020} S.-I.~Mirzadeh \emph{et al.}, ``Improved knowledge distillation via teacher assistant,'' in \emph{Proc. AAAI}, 2020, arXiv:1902.03393.
\bibitem{chen2024overthink} X.~Chen \emph{et al.}, ``Do not think that much for 2+3=? On the overthinking of o1-like LLMs,'' arXiv:2412.21187, 2024.
\bibitem{zheng2025cot} T.~Zheng \emph{et al.}, ``The curse of CoT: On the limitations of chain-of-thought in in-context learning,'' arXiv:2504.05081, 2025.
\bibitem{aggarwal2025} P.~Aggarwal \emph{et al.}, ``OptimalThinkingBench: Evaluating over and underthinking in LLMs,'' arXiv:2508.13141, 2025.
\bibitem{yao2025} Z.~Yao \emph{et al.}, ``Are reasoning models more prone to hallucination?,'' arXiv:2505.23646, 2025.
\bibitem{hsieh2023} C.-Y.~Hsieh \emph{et al.}, ``Distilling step-by-step! Outperforming larger language models with less training data and smaller model sizes,'' in \emph{Findings of ACL}, 2023.
\bibitem{wang2023selfinstruct} Y.~Wang \emph{et al.}, ``Self-Instruct: Aligning language models with self-generated instructions,'' in \emph{Proc. ACL}, 2023, arXiv:2212.10560.
\bibitem{shirgaonkar2024} A.~Shirgaonkar \emph{et al.}, ``Knowledge distillation using frontier open-source LLMs: Generalizability and the role of synthetic data,'' arXiv:2410.18588, 2024.
\bibitem{chrysostomou2023} G.~Chrysostomou \emph{et al.}, ``Investigating hallucinations in pruned large language models for abstractive summarization,'' arXiv:2311.09335, 2023.
\bibitem{ramprasad2024} S.~Ramprasad \emph{et al.}, ``Evaluating the factuality of zero-shot summarizers across varied domains,'' arXiv:2402.03509, 2024.
\bibitem{zhang2023newssumm} T.~Zhang \emph{et al.}, ``Benchmarking large language models for news summarization,'' arXiv:2301.13848, 2023.
\bibitem{song2025ondevice} Q.~Song \emph{et al.}, ``A systematic evaluation of on-device large language models: Quantization, performance, and resources,'' arXiv:2505.15030, 2025.
\bibitem{lin2023awq} J.~Lin \emph{et al.}, ``AWQ: Activation-aware weight quantization for LLM compression and acceleration,'' arXiv:2306.00978, 2023.
\bibitem{zheng2023mtbench} L.~Zheng \emph{et al.}, ``Judging LLM-as-a-judge with MT-Bench and Chatbot Arena,'' arXiv:2306.05685, 2023.
\bibitem{wang2023unfair} P.~Wang \emph{et al.}, ``Large language models are not fair evaluators,'' arXiv:2305.17926, 2023.
\bibitem{ye2024bias} J.~Ye \emph{et al.}, ``Justice or prejudice? Quantifying biases in LLM-as-a-judge,'' arXiv:2410.02736, 2024.
\bibitem{liu2023} Y.~Liu \emph{et al.}, ``G-Eval: NLG evaluation using GPT-4 with better human alignment,'' arXiv:2303.16634, 2023.
\bibitem{lee2024} Y.~Lee \emph{et al.}, ``CheckEval: A reliable LLM-as-a-judge framework for evaluating text generation using checklists,'' arXiv:2403.18771, 2024.
\bibitem{cook2024tick} J.~Cook \emph{et al.}, ``TICKing all the boxes: Generated checklists improve LLM evaluation and generation,'' arXiv:2410.03608, 2024.
\bibitem{qin2024infobench} Y.~Qin \emph{et al.}, ``InFoBench: Evaluating instruction following ability in large language models,'' in \emph{Findings of ACL}, 2024, arXiv:2401.03601.
\bibitem{verga2024poll} P.~Verga \emph{et al.}, ``Replacing judges with juries: Evaluating LLM generations with a panel of diverse models,'' arXiv:2404.18796, 2024.
\bibitem{tian2025overconfidence} Z.~Tian \emph{et al.}, ``Overconfidence in LLM-as-a-judge: Diagnosis and confidence-driven solution,'' arXiv:2508.06225, 2025.
\bibitem{panickssery2024} A.~Panickssery, S.~R.~Bowman, and S.~Feng, ``LLM evaluators recognize and favor their own generations,'' arXiv:2404.13076, 2024.
\bibitem{wataoka2024} K.~Wataoka \emph{et al.}, ``Self-preference bias in LLM-as-a-judge,'' arXiv:2410.21819, 2024.
\bibitem{min2023} S.~Min \emph{et al.}, ``FActScore: Fine-grained atomic evaluation of factual precision in long form text generation,'' arXiv:2305.14251, 2023.
\bibitem{tang2024} L.~Tang, P.~Laban, and G.~Durrett, ``MiniCheck: Efficient fact-checking of LLMs on grounding documents,'' arXiv:2404.10774, 2024.
\bibitem{jiang2023tigerscore} D.~Jiang \emph{et al.}, ``TIGERScore: Towards building explainable metric for all text generation tasks,'' arXiv:2310.00752, 2023.
\bibitem{manakul2023} P.~Manakul, A.~Liusie, and M.~J.~F.~Gales, ``SelfCheckGPT: Zero-resource black-box hallucination detection for generative large language models,'' arXiv:2303.08896, 2023.
\bibitem{geng2025jsonschema} S.~Geng \emph{et al.}, ``JSONSchemaBench: A rigorous benchmark of structured outputs for language models,'' arXiv:2501.10868, 2025.
\end{thebibliography}
\end{document}